%% file: main.tex
\documentclass[11pt,letterpaper]{article}
\usepackage{emnlp2016}
\usepackage{times}
\usepackage{latexsym}

\usepackage{epsfig}
\usepackage{graphicx}
\usepackage[belowskip=-10pt,aboveskip=2pt,font=small]{caption}

\usepackage{amsmath}
\usepackage{amssymb}

\usepackage{multirow}
\usepackage{rotating}
\usepackage{booktabs}
\usepackage{slashbox} 
\usepackage{algpseudocode}
\usepackage{algorithm}
\usepackage{xspace}
\usepackage{enumerate}
\usepackage[olditem,oldenum]{paralist}
\usepackage{soul}

\input{space_saver}

\usepackage{color}
\usepackage{comment}
\usepackage[pagebackref=true,breaklinks=true,letterpaper=true,colorlinks,bookmarks=false,citecolor=blue,linkcolor=green]{hyperref}

\emnlpfinalcopy

\def\emnlppaperid{26}

\newcommand{\noatt}{\textsc{CNN+LSTM}\xspace}
\newcommand{\att}{\textsc{ATT}\xspace}
\newcommand{\figref}[1]{Fig.~\ref{#1}}

\newcommand{\mcb}{\textsc{MCB}\xspace}

\title{Analyzing the Behavior of Visual Question Answering Models}
\author{Aishwarya Agrawal$^{\ast}$, Dhruv Batra$^{\dagger, \ast}$, Devi Parikh$^{\dagger, \ast}$ \\
$^{\ast}$Virginia Tech \quad $^{\dagger}$Georgia Institute of Technology\\
{\tt\{aish, dbatra, parikh\}@vt.edu}
}

\begin{document}

\maketitle

\begin{abstract}
Recently, a number of deep-learning based models have been proposed for the task of Visual Question Answering (VQA). The performance of most models is clustered around 60-70\%. In this paper we propose systematic methods to analyze the behavior of these models as a first step towards recognizing their strengths and weaknesses, and identifying the most fruitful directions for progress. We analyze two models, one each from two major classes of VQA models -- with-attention and without-attention and show the similarities and differences in the behavior of these models. We also analyze the winning entry of the VQA Challenge 2016.

Our behavior analysis reveals that despite recent progress, today's VQA models are ``myopic'' (tend to fail on sufficiently novel instances), often ``jump to conclusions'' (converge on a predicted answer after `listening' to just half the question), and are ``stubborn'' (do not change their answers across images).
\end{abstract}

\section{Introduction}
Visual Question Answering (VQA) is a recently-introduced \cite{VQA,geman,fritz} problem where given an image and a natural language question (e.g., ``What kind of store is this?'', ``How many people are waiting in the queue?''), the task is to automatically produce an accurate natural language answer (``bakery'', ``5''). A flurry of recent deep-learning based models have been proposed for VQA \cite{VQA,ChenWCGXN15,DBLP:journals/corr/YangHGDS15,DBLP:journals/corr/XuS15a,DBLP:journals/corr/JiangWPL15,DBLP:journals/corr/AndreasRDK15,DBLP:journals/corr/WangWSHD15,kanan,hieco,DBLP:journals/corr/AndreasRDK16,DBLP:journals/corr/ShihSH15,DBLP:journals/corr/kim15,fukui,han,DBLP:journals/corr/IlievskiYF16,DBLP:journals/corr/WuWSHD15,DBLP:journals/corr/XiongMS16,zhou,saito}. Curiously, the performance of most methods is clustered around 60-70\% (compared to human performance of 83\% on open-ended task and 91\% on multiple-choice task) with a mere 5\% gap between the top-9 entries on the VQA Challenge 2016.\footnote{\url{http://www.visualqa.org/challenge.html}} It seems clear that as a first step to understand these models, to meaningfully compare strengths and weaknesses of different models, to develop insights into their failure modes, and to identify the most fruitful directions for progress, it is crucial to develop techniques to understand the behavior of VQA models. 

In this paper, we develop novel techniques to characterize the behavior of VQA models. As concrete instantiations, we analyze two VQA models \cite{Lu2015,hieco}, one each from two major classes of VQA models -- with-attention and without-attention. We also analyze the winning entry \cite{fukui} of the VQA Challenge 2016.

\section{Related Work}
Our work is inspired by previous works that diagnose the failure modes of models for different tasks. \cite{DBLP:journals/corr/KarpathyJL15} constructed a series of oracles to measure the performance of a character level language model. \cite{Hoiem} provided analysis tools to facilitate detailed and meaningful investigation of object detector performance. This paper aims to perform behavior analyses as a first step towards diagnosing errors for VQA.

\cite{DBLP:journals/corr/YangHGDS15} categorize the errors made by their VQA model into four categories -- model focuses attention on incorrect regions, model focuses attention on appropriate regions but predicts incorrect answers, predicted answers are different from labels but might be acceptable, labels are wrong. While these are coarse but useful failure modes, we are interested in understanding the behavior of VQA models along specific dimensions -- whether they generalize to novel instances, whether they listen to the entire question, whether they look at the image.


\section{Behavior Analyses}
We analyze the behavior of VQA models along the following three dimensions --

\textbf{Generalization to novel instances:} We investigate whether the test instances that are incorrectly answered are the ones that are ``novel'' i.e., not similar to training instances. The novelty of the test instances may be in two ways --  1) the test question-image (QI) pair is ``novel'', i.e., too different from training QI pairs; and 2) the test QI pair is ``familiar'', but the answer required at test time is ``novel'', i.e., answers seen during training are different from what needs to be produced for the test QI pairs. 

\textbf{Complete question understanding:} To investigate whether a VQA model is understanding the input question or not, we analyze if the model `listens' to only first few words of the question or the entire question, if it `listens' to only question (wh) words and nouns or all the words in the question. 

\textbf{Complete image understanding:} The absence of a large gap between performance of language-alone and language + vision VQA models \cite{VQA} provides evidence that current VQA models seem to be heavily reliant on the language model, perhaps not really understanding the image. In order to analyze this behavior, we investigate whether the predictions of the model change across images for a given question. 

We present our behavioral analyses on the VQA dataset \cite{VQA}. VQA is a large-scale free-form natural-language dataset containing $\sim$0.25M images, $\sim$0.76M questions, and $\sim$10M answers, with open-ended and multiple-choice modalities for answering the visual questions. All the experimental results are reported on the VQA validation set using the following models trained on the VQA training set for the open-ended task -- 

\textbf{CNN + LSTM based model without-attention (\noatt):} We use the best performing model of \cite{VQA} (code provided by \cite{Lu2015}), which achieves an accuracy of 54.13\% on the VQA validation set. It is a two channel model -- one channel processes the image (using Convolutional Neural Network (CNN) to extract image features) and the other channel processes the question (using Long Short-Term Memory (LSTM) recurrent neural network to obtain question embedding). The image and question features obtained from the two channels are combined and passed through a fully connected (FC) layer to obtain a softmax distribution over the space of answers. 


\textbf{CNN + LSTM based model with-attention (\att):} We use the top-entry on the VQA challenge leaderboard (as of June 03, 2016) \cite{hieco}, which achieves an accuracy of 57.02\% on the VQA validation set.\footnote{Code available at \url{https://github.com/jiasenlu/HieCoAttenVQA}} This model jointly reasons about image and question attention, in a hierarchical fashion. The attended image and question features obtained from different levels of the hierarchy are combined and passed through a FC layer to obtain a softmax distribution over the space of answers. 

\textbf{VQA Challenge 2016 winning entry (\mcb):} This is the multimodal compact bilinear (mcb) pooling model from \cite{fukui} which won the real image track of the VQA Challenge 2016. This model achieves an accuracy of 60.36\% on the VQA validation set.\footnote{Code available at \url{https://github.com/akirafukui/vqa-mcb}} In this model, multimodal compact bilinear pooling is used to predict attention over image features and also to combine the attended image features with the question features. These combined features are passed through a FC layer to obtain a softmax distribution over the space of answers.

\subsection{Generalization to novel instances}
Do VQA models make mistakes because test instances are too different from training ones? To analyze the first type of novelty (the test QI pair is novel), we measure the correlation between test accuracy and distance of test QI pairs from its k nearest neighbor (k-NN) training QI pairs. For each test QI pair we find its k-NNs in the training set and compute the average distance between the test QI pair and its k-NNs. The k-NNs are computed in the space of combined image + question embedding (just before passing through FC layer) for all the three models (using euclidean distance metric for the \noatt model and cosine distance metric for the \att and \mcb models).

%

The correlation between accuracy and average distance is significant (-0.41 at k=50\footnote{k=50 leads to highest correlation} for the \noatt model and -0.42 at k=15\footnote{k=15 leads to highest correlation} for the \att model). A high negative correlation value tells that the model is less likely to predict correct answers for test QI pairs which are not very similar to training QI pairs, suggesting that the model is not very good at generalizing to novel test QI pairs. The correlation between accuracy and average distance is not significant for the \mcb model (-0.14 at k=1\footnote{k=1 leads to highest correlation}) suggesting that \mcb is better at generalizing to novel test QI pairs.

We also found that 67.5\% of mistakes made by the \noatt model \emph{can be successfully predicted} by checking distance of test QI pair from its k-NN training QI pairs (66.7\% for the \att model, 55.08\% for the \mcb model). Thus, this analysis not only exposes a reason for mistakes made by VQA models, but also allows us to build human-like models that can predict their own oncoming failures, and potentially refuse to answer questions that are `too different' from ones seen in past. 

To analyze the second type of novelty (the answer required at test time is not familiar), we compute the correlation between test accuracy and the average distance of the test ground truth (GT) answer with GT answers of its k-NN training QI pairs. The distance between answers is  computed in the space of average Word2Vec \cite{mikolov2013efficient} vectors of answers. This correlation turns out to be quite high (-0.62) for both \noatt and \att models and significant (-0.47) for the \mcb model. A high negative correlation value tells that the model tends to regurgitate answers seen during training. 

\begin{figure}[t]
\centering
\includegraphics[width=1\linewidth]{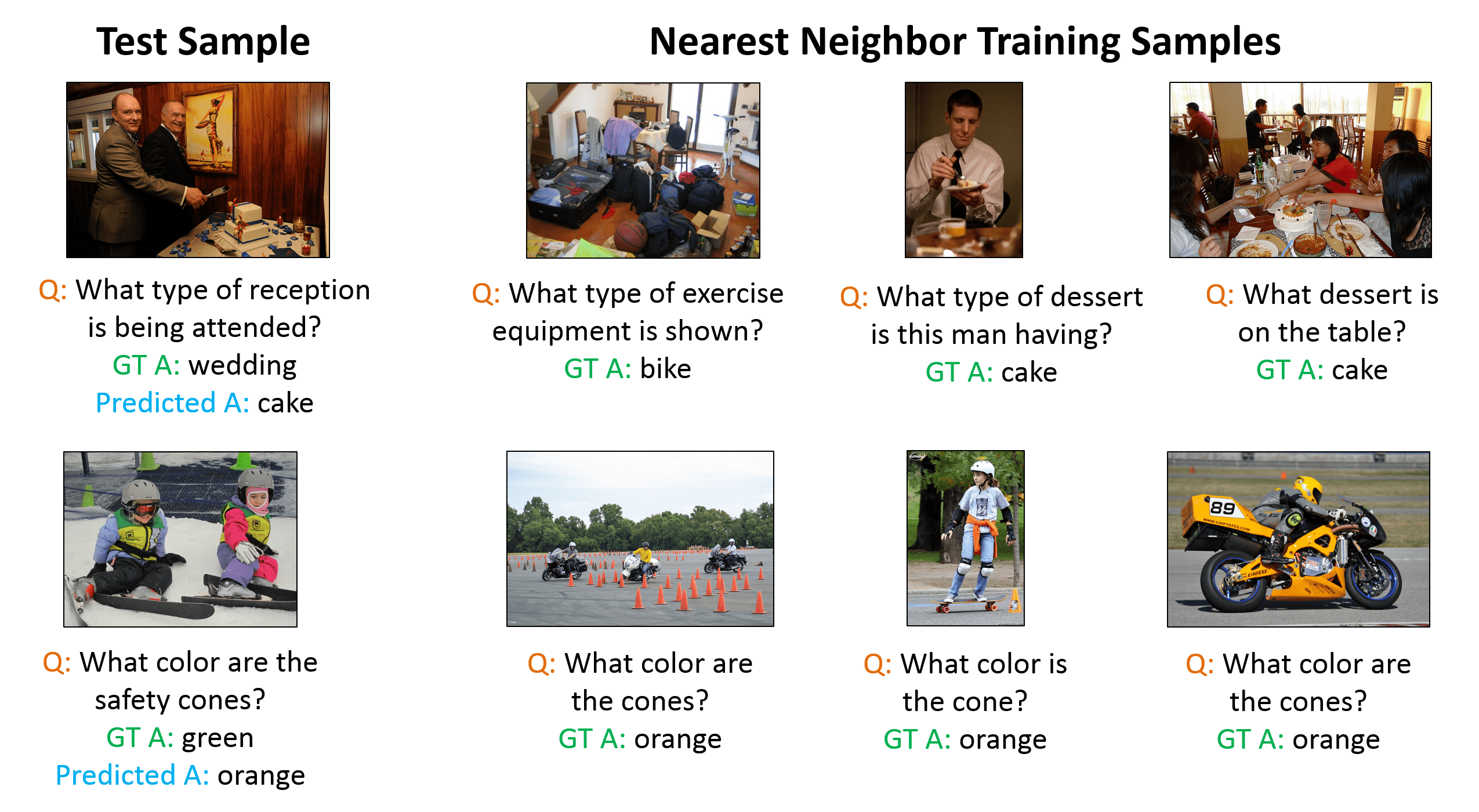}
\caption{Examples from test set where the \noatt model makes mistakes and their corresponding nearest neighbor training instances. See \hyperref[sec:sec3]{Appendix III} for more examples.}
\label{fig:qual1}
\end{figure}

These distance features are also good at predicting failures --  74.19\% of failures can be predicted by checking distance of test GT answer with GT answers of its k-NN training QI pairs for \noatt model (75.41\% for the \att model, 70.17\% for the \mcb model). Note that unlike the previous analysis, this analysis only explains failures but cannot be used to predict failures (since it uses GT labels). See \figref{fig:qual1} for qualitative examples.


From \figref{fig:qual1} (row1) we can see that the test QI pair is semantically quite different from its k-NN training QI pairs (\{1st, 2nd, 3rd\}-NN distances are \{15.05, 15.13, 15.17\}, which are higher than the corresponding distances averaged across all success cases: \{8.74, 9.23, 9.50.\}), explaining the mistake. Row2 shows an example where the model has seen the same question in the training set (test QI pair is semantically similar to training QI pairs) but, since it has not seen ``green cone'' for training instances (answers seen during training are different from what needs to be produced for the test QI pair), it is unable to answer the test QI pair correctly. This shows that current models lack compositionality: the ability to combine the concepts of ``cone'' and ``green'' (both of which have been seen in training set) to answer ``green cone'' for the test QI pair. This compositionality is desirable and central to intelligence.
%

\subsection{Complete question understanding}
We feed partial questions of increasing lengths (from 0-100\% of question from left to right). We then compute what percentage of responses do not change when more and more words are fed. 

\begin{figure}[t]
\centering
\includegraphics[width=1\linewidth]{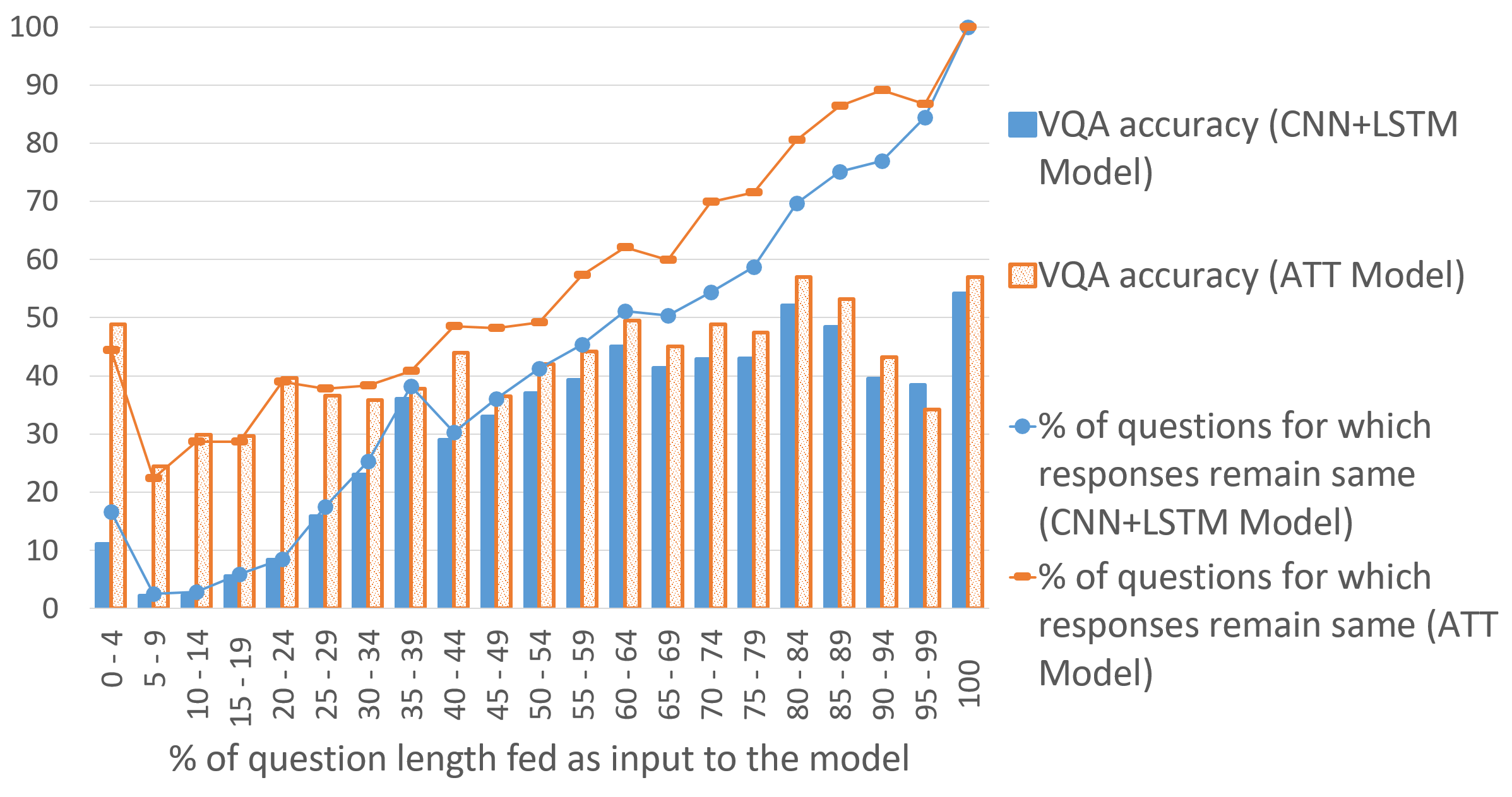}
\caption{X-axis shows length of partial question (in \%) fed as input. Y-axis shows percentage of questions for which responses of these partial questions are the same as full questions and VQA accuracy of partial questions.}
\label{fig:percent_ques_len_combined}
\end{figure}

\begin{figure}[h]
\centering
\includegraphics[width=1\linewidth]{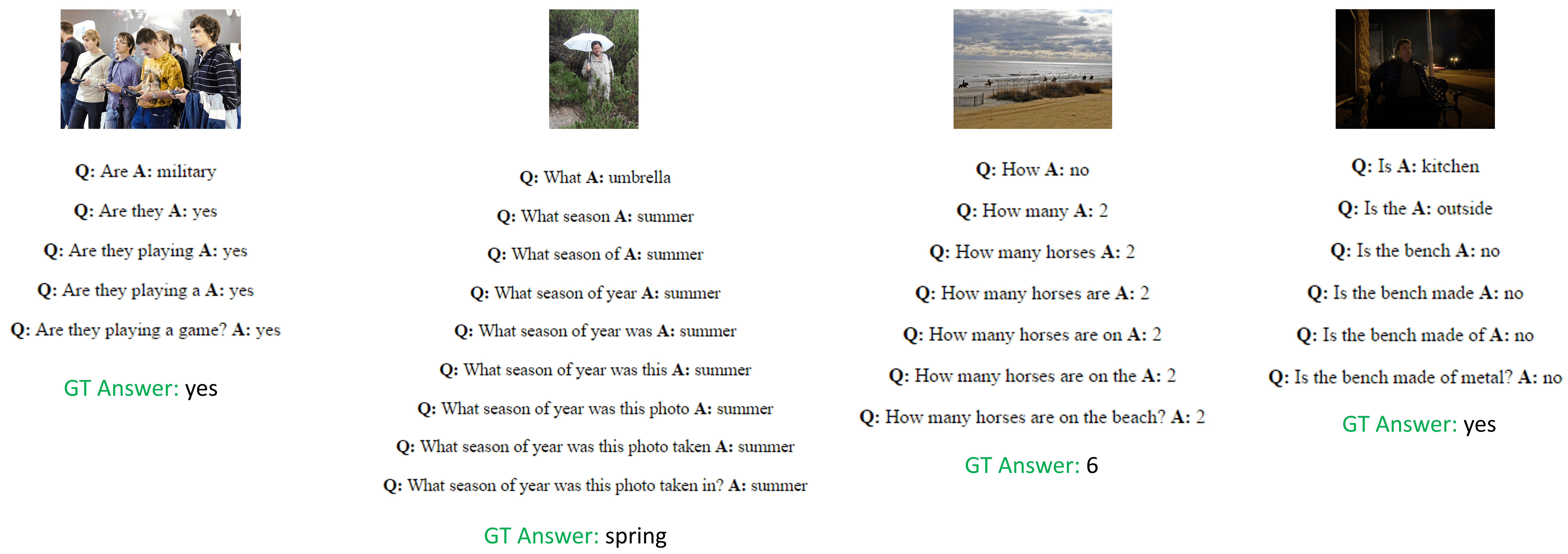}
\caption{Examples where the \noatt model does not change its answer after first few question words. On doing so, it is correct for some cases (the extreme left example) and incorrect for other cases (the remaining three examples). See \hyperref[sec:sec5]{Appendix V} for more examples.}
\label{fig:qual2}
\end{figure}

\figref{fig:percent_ques_len_combined} shows the test accuracy and percentage of questions for which responses remain same (compared to entire question) as a function of partial question length. We can see that for 40\% of the questions, the \noatt model seems to have converged on a predicted answer after `listening' to just half the question. This shows that the model is listening to first few words of the question more than the words towards the end. Also, the model has 68\% of the final accuracy (54\%) when making predictions based on half the original question. When making predictions just based on the image, the accuracy of the model is 24\%. The \att model seems to have converged on a predicted answer after listening to just half the question more often (49\% of the time), achieving 74\% of the final accuracy (57\%). The \mcb model converges on a predicted answer after listening to just half the question 45\% of the time, achieving 67\% of the final accuracy (60\%). See \figref{fig:qual2} for qualitative examples.

%

\begin{figure}[h]
\centering
\includegraphics[width=1\linewidth]{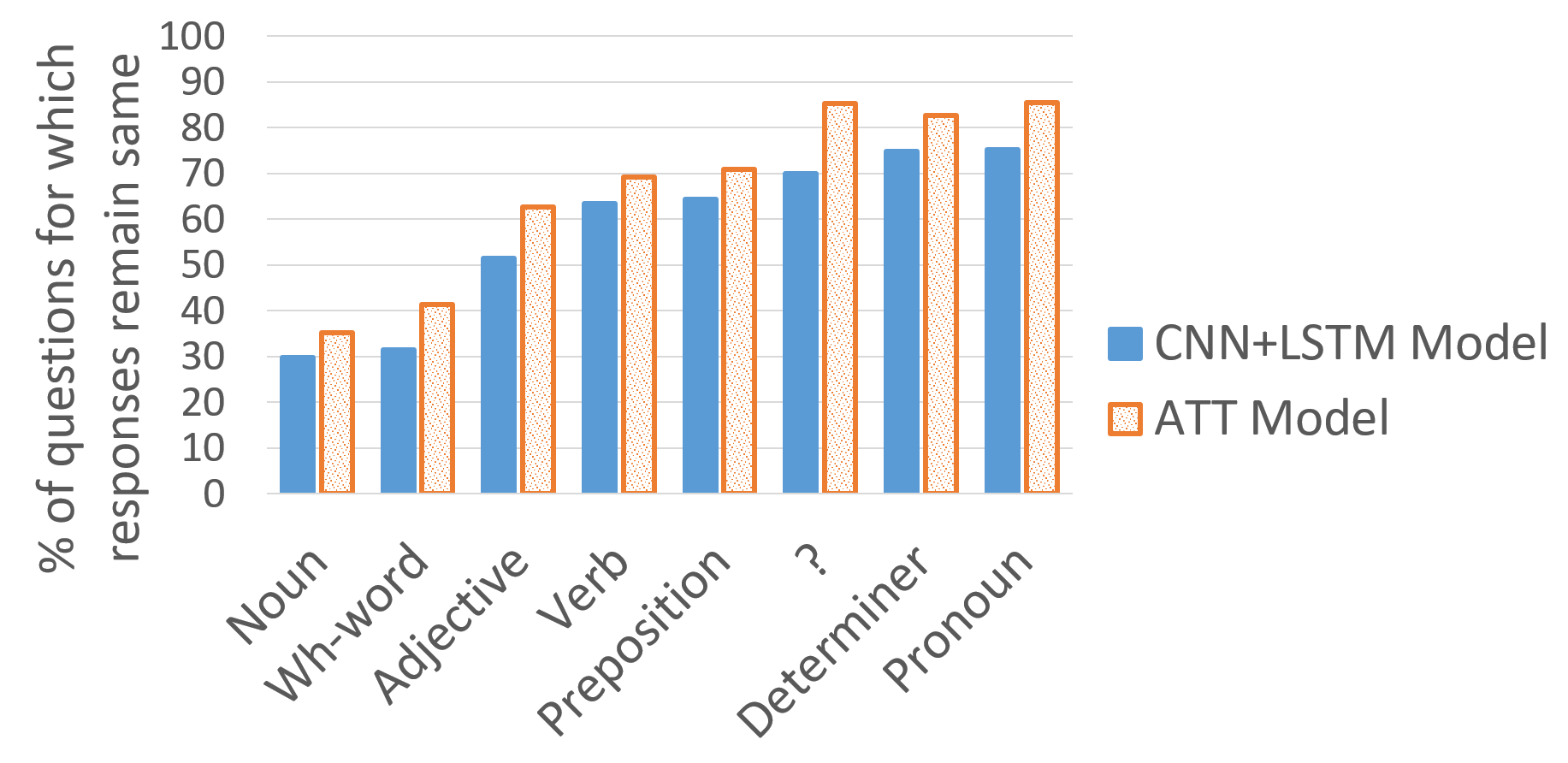}
\caption{Percentage of questions for which responses remain same (compared to entire question) as a function of POS tags dropped from the question.}
\label{fig:pos}
\end{figure}

We also analyze the change in responses of the model's predictions (see \figref{fig:pos}), when words of a particular part-of-the-speech (POS) tag are dropped from the question. The experimental results indicate that wh-words effect the model's decisions the most (most of the responses get changed on dropping these words from the question), and that pronouns effect the model's decisions the least.

%

\subsection{Complete image understanding}
\begin{figure}[h]
\centering
\includegraphics[width=1\linewidth]{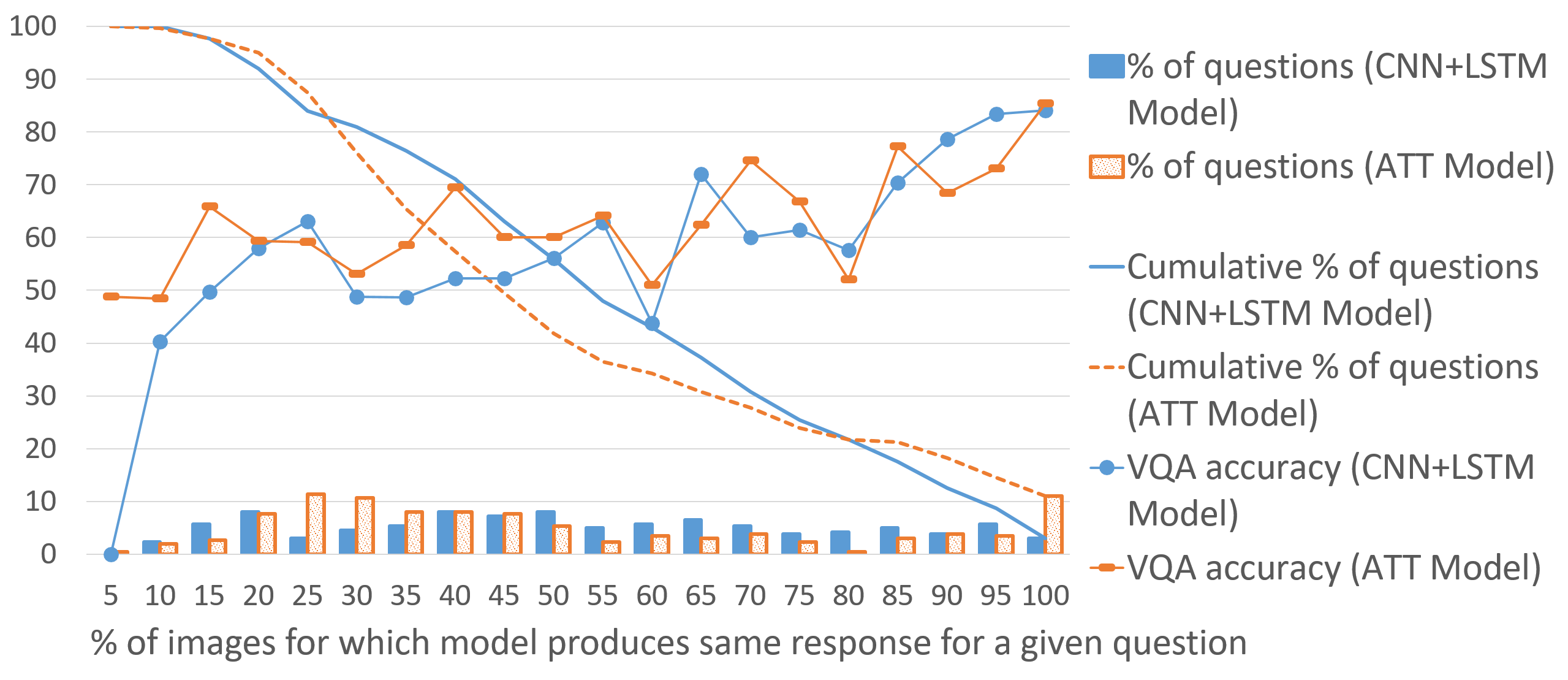}
\caption{Histogram of percentage of images for which model produces same answer for a given question and its comparison with test accuracy. The cumulative plot shows the \% of questions for which model produces same answer for \emph{atleast} $x$ \% of images.}
\label{fig:image_understanding}
\end{figure}

Does a VQA model really `look' at the image? To analyze this, we compute the percentage of the time (say $X$) the response does not change across images (e.g.,, answer for all images is ``2'') for a given question (e.g., ``How many zebras?'') and plot histogram of $X$ across questions (see \figref{fig:image_understanding}). We do this analysis for questions occurring for atleast 25 images in the VQA validation set, resulting in total 263 questions. The cumulative plot indicates that for 56\% questions, the \noatt model outputs the same answer for at least half the images. This is fairly high, suggesting that the model is picking the same answer no matter what the image is. Promisingly, the \att and \mcb models (that do not work with a holistic entire-image representation and purportedly pay attention to specific spatial regions in an image) produce the same response for at least half the images for fewer questions (42\% for the \att model, 40\% for the \mcb model).

\begin{figure}[t]
\centering
\includegraphics[width=1\linewidth]{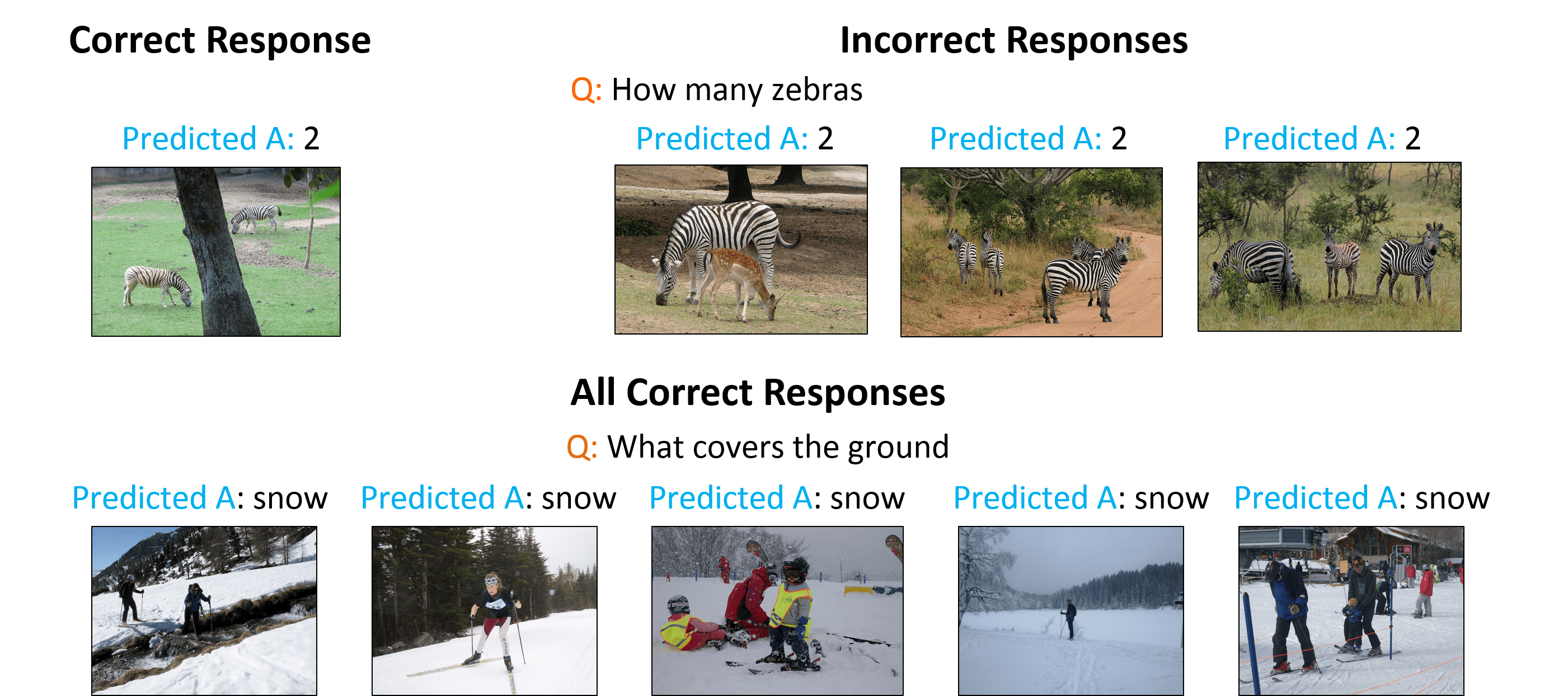}
\caption{Examples where the predicted answers do not change across images for a given question. See \hyperref[sec:sec7]{Appendix VI} for more examples.}
\label{fig:qual3}
\end{figure}

Interestingly, the average accuracy (see the VQA accuracy plots in \figref{fig:image_understanding}) for questions for which the models produce same response for $>$50\% and $<$55\% of the images is 56\% for the \noatt model (60\% for the \att model, 73\% for the \mcb model) which is more than the respective average accuracy on the entire VQA validation set (54.13\% for the \noatt model, 57.02\% for the \att model, 60.36\% for the \mcb model). Thus, producing the same response across images seems to be statistically favorable. \figref{fig:qual3} shows examples where the \noatt model predicts the same response across images for a given question. The first row shows examples where the model makes errors on several images by predicting the same answer for all images. The second row shows examples where the model is always correct even if it predicts the same answer across images. This is so because questions such as \emph{``What covers the ground?''} are asked for an image in the VQA dataset only when ground is covered with snow (because subjects were looking at the image while asking questions about it). Thus, this analysis exposes label biases in the dataset. Label biases (in particular, for ``yes/no'' questions) have also been reported in \cite{YinYang}.

\section{Conclusion}
We develop novel techniques to characterize the behavior of VQA models, as a first step towards understanding these models, meaningfully comparing the strengths and weaknesses of different models, developing insights into their failure modes, and identifying the most fruitful directions for progress. Our behavior analysis reveals that despite recent progress, today's VQA models are ``myopic'' (tend to fail on sufficiently novel instances), often ``jump to conclusions'' (converge on a predicted answer after `listening' to just half the question), and are ``stubborn'' (do not change their answers across images), with attention based models being less ``stubborn'' than non-attention based models.

As a final thought, we note that the somewhat pathological behaviors exposed in the paper are in some sense ``correct'' given the model architectures and the dataset being trained on. Ignoring optimization error, the maximum-likelihood training objective is clearly intended to capture statistics of the dataset. Our motive is simply to better understand current generation models via their behaviors, and use these observations to guide future choices -- do we need novel model classes? or dataset with different biases? etc. Finally, it should be clear that our use of anthropomorphic adjectives such as ``stubborn'', ``myopic'' etc. is purely for pedagogical reasons -- to easily communicate our observations to our readers. No claims are being made about today's VQA models being human-like. 


\textbf{Acknowledgements.}
We would like to thank the EMNLP reviewers for their valuable feedback and Yash Goyal for sharing his code. This work was supported in part by: 
NSF CAREER awards, 
Army Research Office YIP awards, 
ICTAS Junior Faculty awards, 
Google Faculty Research awards, awarded to both DB and DP, 
ONR grant N00014-14-1-0679,
AWS in Education Research grant,
NVIDIA GPU donation, awarded to DB, 
Paul G. Allen Family Foundation Allen Distinguished Investigator award, 
ONR YIP 
and 
Alfred P. Sloan Fellowship, awarded to DP.
The views and conclusions contained herein are those of the authors and should not be interpreted as 
necessarily representing the official policies or endorsements, either expressed or implied, of the U.S.
Government or any sponsor. 

\input{supp}

\bibliography{emnlp2016}
\bibliographystyle{emnlp2016}

\end{document}

%% file: space_saver.tex
\belowdisplayskip \abovedisplayskip

\newlength{\sectionReduceTop}
\newlength{\sectionReduceBot}
\newlength{\subsectionReduceTop}
\newlength{\subsectionReduceBot}
\newlength{\abstractReduceTop}
\newlength{\abstractReduceBot}
\newlength{\captionReduceTop}
\newlength{\captionReduceBot}
\newlength{\subsubsectionReduceTop}
\newlength{\subsubsectionReduceBot}

\newlength{\eqnReduceTop}
\newlength{\eqnReduceBot}

\newlength{\horSkip}
\newlength{\verSkip}

\newlength{\figureHeight}

%% file: supp.tex
\section*{Appendix Overview}
In the appendix, we provide:
\begin{enumerate}[I]
\setlength{\itemsep}{1pt}
  \setlength{\parskip}{0pt}
  \setlength{\parsep}{0pt}
\item - Behavioral analysis for question-only and image-only VQA models (\hyperref[sec:sec1]{Appendix I}).
\item - Scatter plot of average distance of test instances from nearest neighbor training instances w.r.t. VQA accuracy (\hyperref[sec:sec2]{Appendix II}).
\item - Additional qualitative examples for ``generalization to novel test instances'' (\hyperref[sec:sec3]{Appendix III}).
\item - The analyses on ``complete question understanding'' for different question types (\hyperref[sec:sec4]{Appendix IV}).
\item - Additional qualitative examples for ``complete question understanding'' (\hyperref[sec:sec5]{Appendix V}).
\item - The analyses on ``complete image understanding'' for different question types (\hyperref[sec:sec6]{Appendix VI}).
\item - Additional qualitative examples for ``complete image understanding'' (\hyperref[sec:sec7]{Appendix VII}).
\end{enumerate}

\section*{Appendix I: Behavioral analysis for question-only and image-only VQA models}
\label{sec:sec1}
We evaluated the performance of both \noatt and \att models by just feeding in the question (and mean image embedding) and by just feeding in the image (and mean question embedding). We computed the percentage of responses that change on feeding the question as well, compared to only feeding in the image and the percentage of responses that change on feeding the image as well, compared to only feeding in the question. We found that that the responses changed much more (about 40\% more) on addition of the question than they did on addition of the image. So this suggests that the VQA models are heavily driven by question rather than the image.


\section*{Appendix II: Scatter plot of average distance of test instances from nearest neighbor training instances w.r.t. VQA accuracy }
\label{sec:sec2}

\begin{figure}[h]
\centering
\includegraphics[width=1\linewidth]{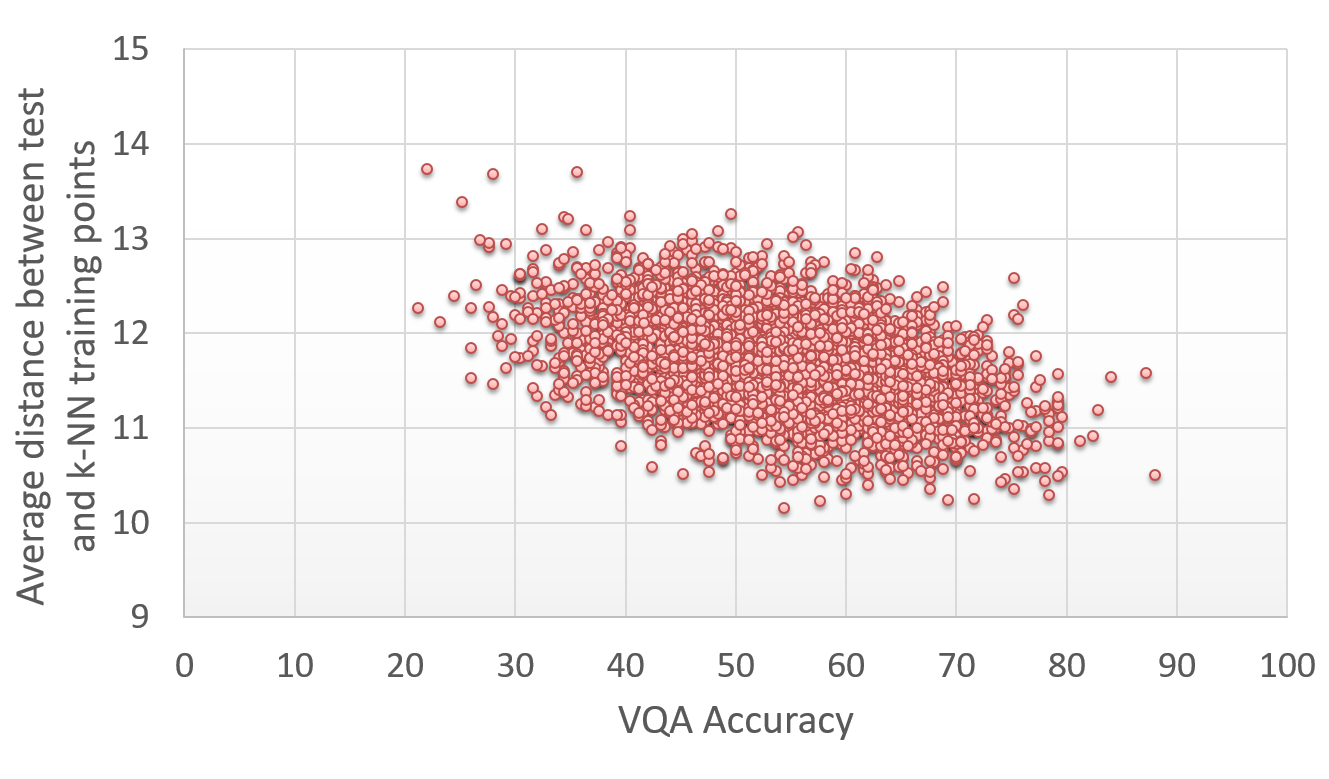}
\caption{Test accuracy vs. average distance of the test points from k-NN training points for the \noatt model.}
\label{fig:acc_dist_corr}
\end{figure}

\figref{fig:acc_dist_corr} shows the variation of accuracy of test point w.r.t their average distance from k-NN training points for the \noatt model. Each point in the plot represents average statistics (accuracy and average distance) for a random subset of 25 test points. We can see that for the test points with low accuracy, the average distance is higher compared to test points with high accuracy. The correlation between accuracy and average distance is significant (-0.41 at $k = 50$.\footnote{$k=50$ leads to highest correlation}) 

\section*{Appendix III: Additional qualitative examples for ``generalization to novel test instances''}
\label{sec:sec3}

\begin{figure*}[h]
\centering
\includegraphics[width=1\linewidth]{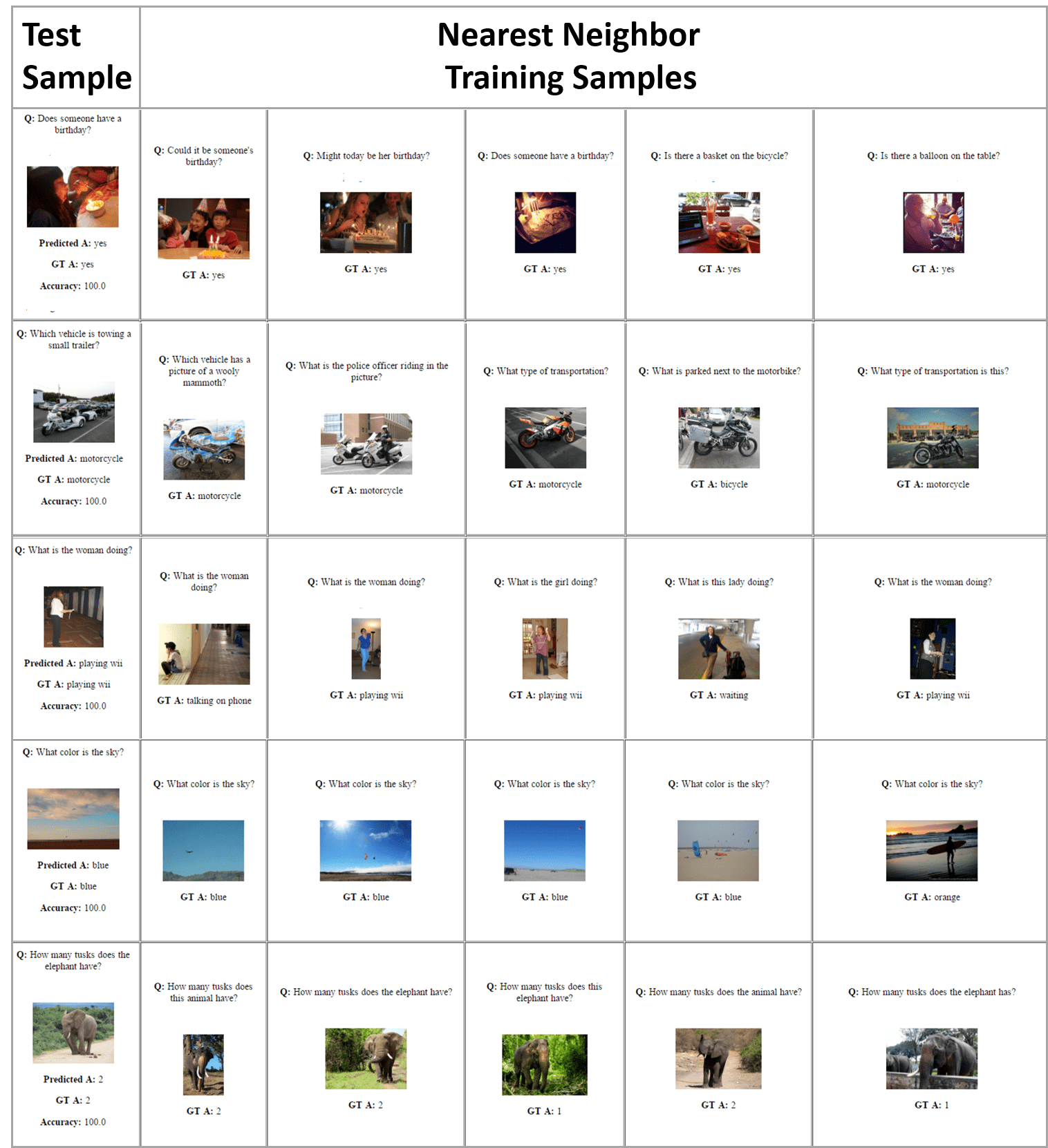}
\caption{Test QI pairs for which the \noatt model produces the correct response and their nearest neighbor QI pairs from training set.}
\label{fig:qual1_supp}
\end{figure*}

\figref{fig:qual1_supp} shows test QI pairs for which the \noatt model produces the correct response and their nearest neighbor QI pairs from training set. It can be seen that the nearest neighbor QI pairs from the training set are similar to the test QI pair. In addition, the GT labels in the training set are similar to the test GT label.

\begin{figure*}[h]
\centering
\includegraphics[width=1\linewidth]{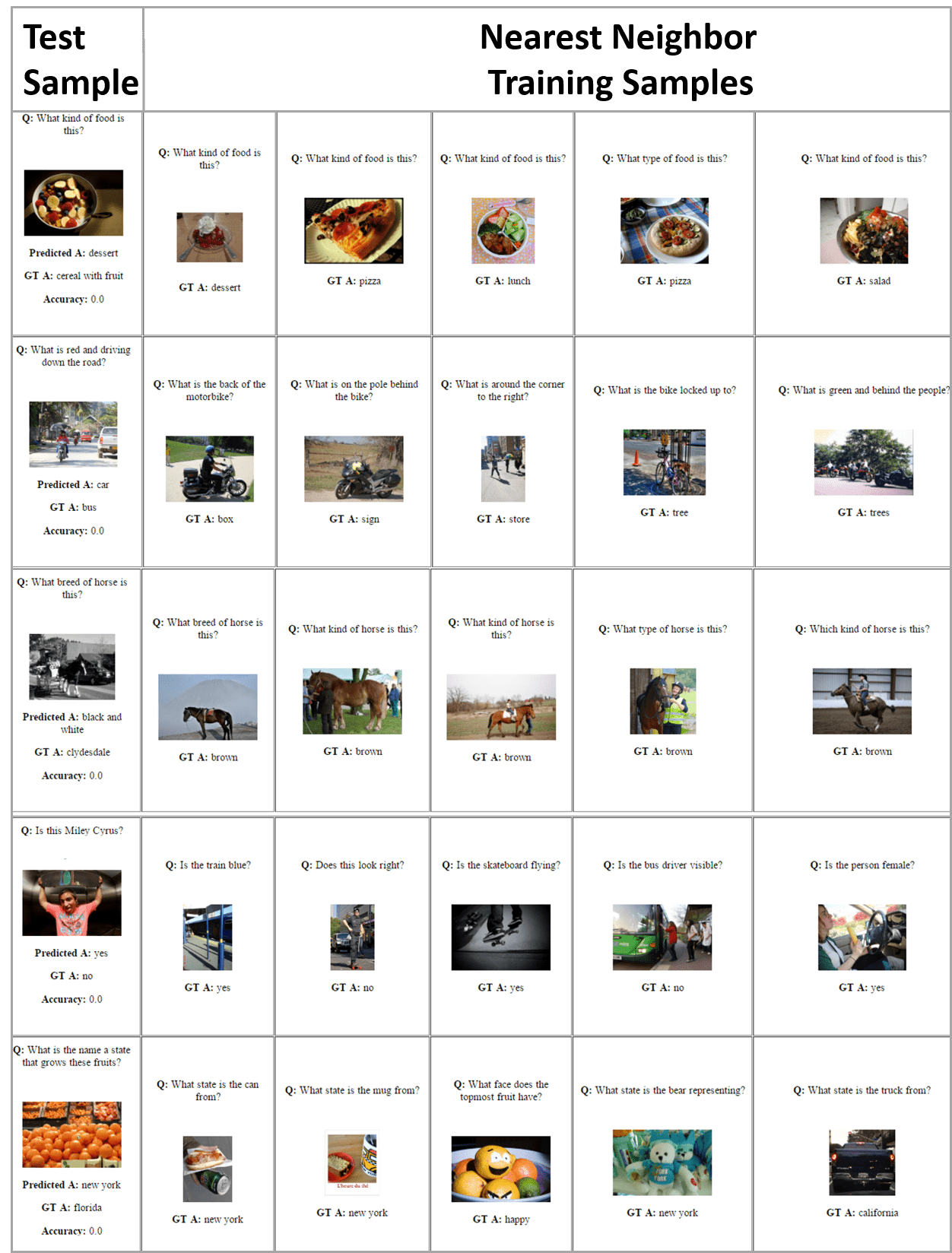}
\caption{Test QI pairs for which the \noatt model produces incorrect response and their nearest neighbor QI pairs from training set.}
\label{fig:qual2_supp}
\end{figure*}

\figref{fig:qual2_supp} shows test QI pairs for which the \noatt model produces incorrect response and their nearest neighbor QI pairs from training set. Some of the mistakes are probably because the test QI pair does not have similar QI pairs in the training set (rows 2, 4 and 5) while other mistakes are probably because the GT labels in the training set are not similar to the GT test label (rows 1 and 3).

\section*{Appendix IV: Analyses on ``complete question understanding'' for different question types}
\label{sec:sec4}

We show the breakdown of our analyses from the main paper -- (i) whether the model `listens' to the entire question; and (ii) which POS tags matter the most -- over the three major categories of questions -- ``yes/no'', ``number'' and ``other'' as categorized in \cite{VQA}. ``yes/no'' are questions whose answers are either ``yes'' or ``no'', ``number'' are questions whose answers are numbers (e.g., ``Q: How many zebras are there?'', ``A: 2''), ``other'' are rest of the questions.

\begin{figure}[h]
\centering
\includegraphics[width=1\linewidth]{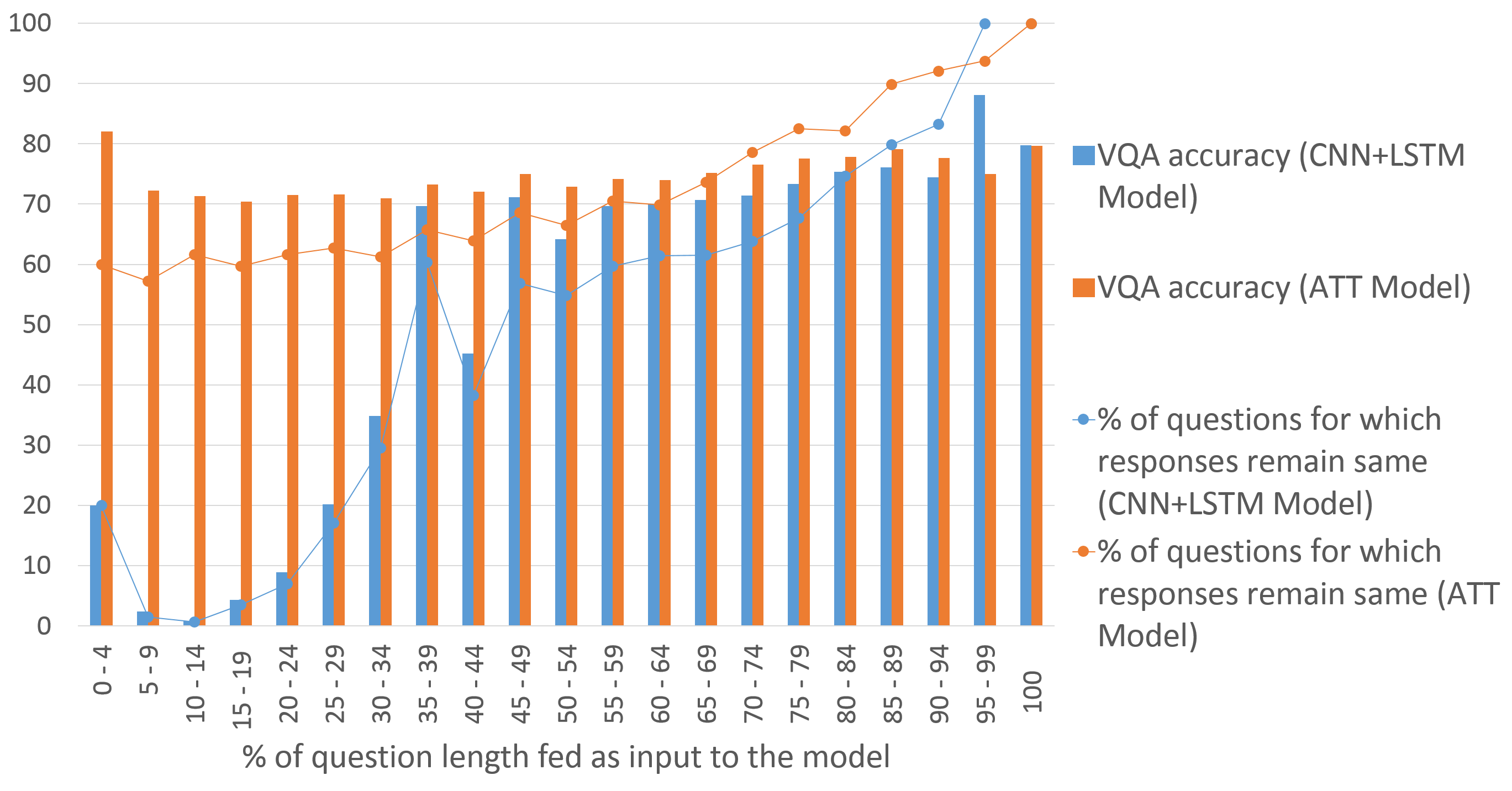}
\caption{X-axis shows length of partial ``yes/no'' question (in \%) fed as input. Y-axis shows percentage of ``yes/no'' questions for which responses of these partial ``yes/no'' questions are the same as full ``yes/no'' questions and VQA accuracy of partial ``yes/no'' questions.}
\label{fig:ques_len_bin}
\end{figure}

\begin{figure}[h]
\centering
\includegraphics[width=1\linewidth]{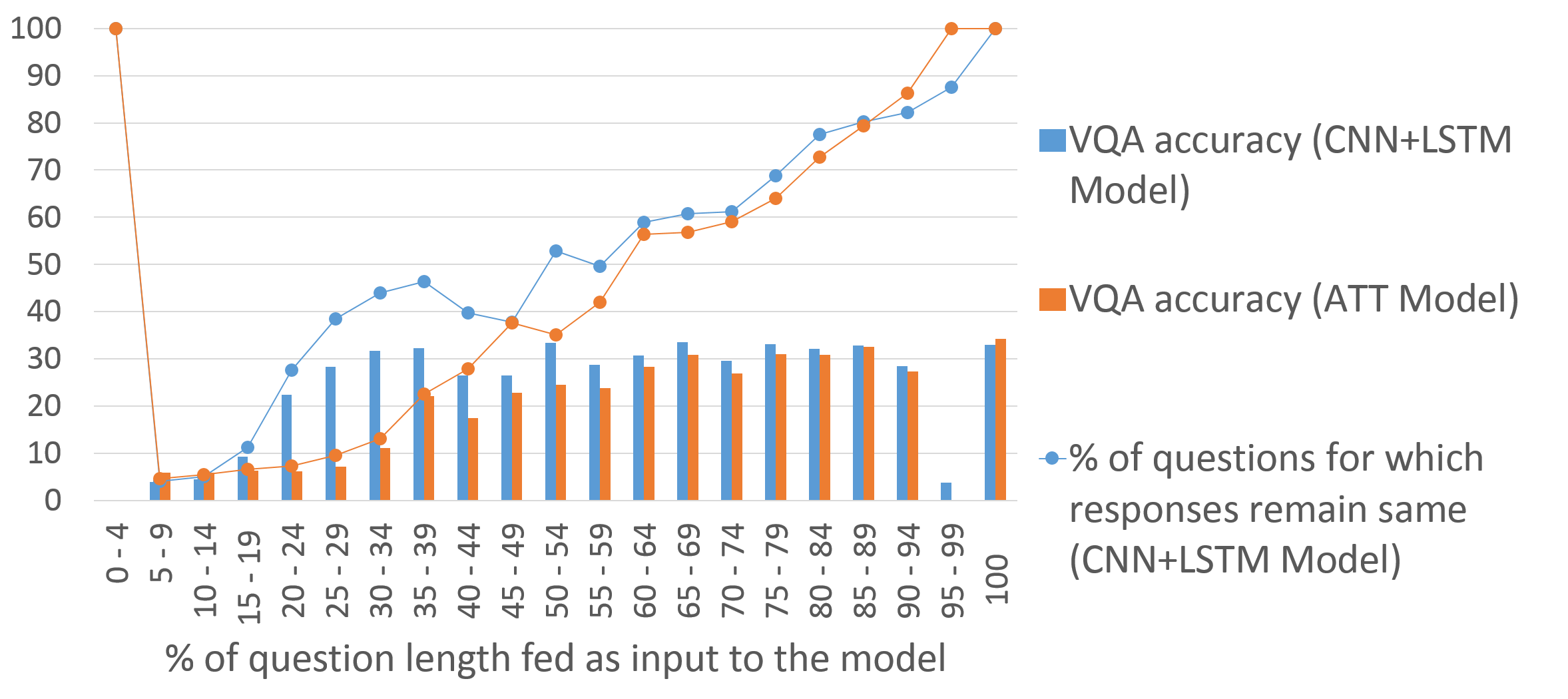}
\caption{X-axis shows length of partial ``number'' question (in \%) fed as input. Y-axis shows percentage of ``number'' questions for which responses of these partial ``number'' questions are the same as full ``number'' questions and VQA accuracy of partial ``number'' questions.}
\label{fig:ques_len_num}
\end{figure}

\begin{figure}[h]
\centering
\includegraphics[width=1\linewidth]{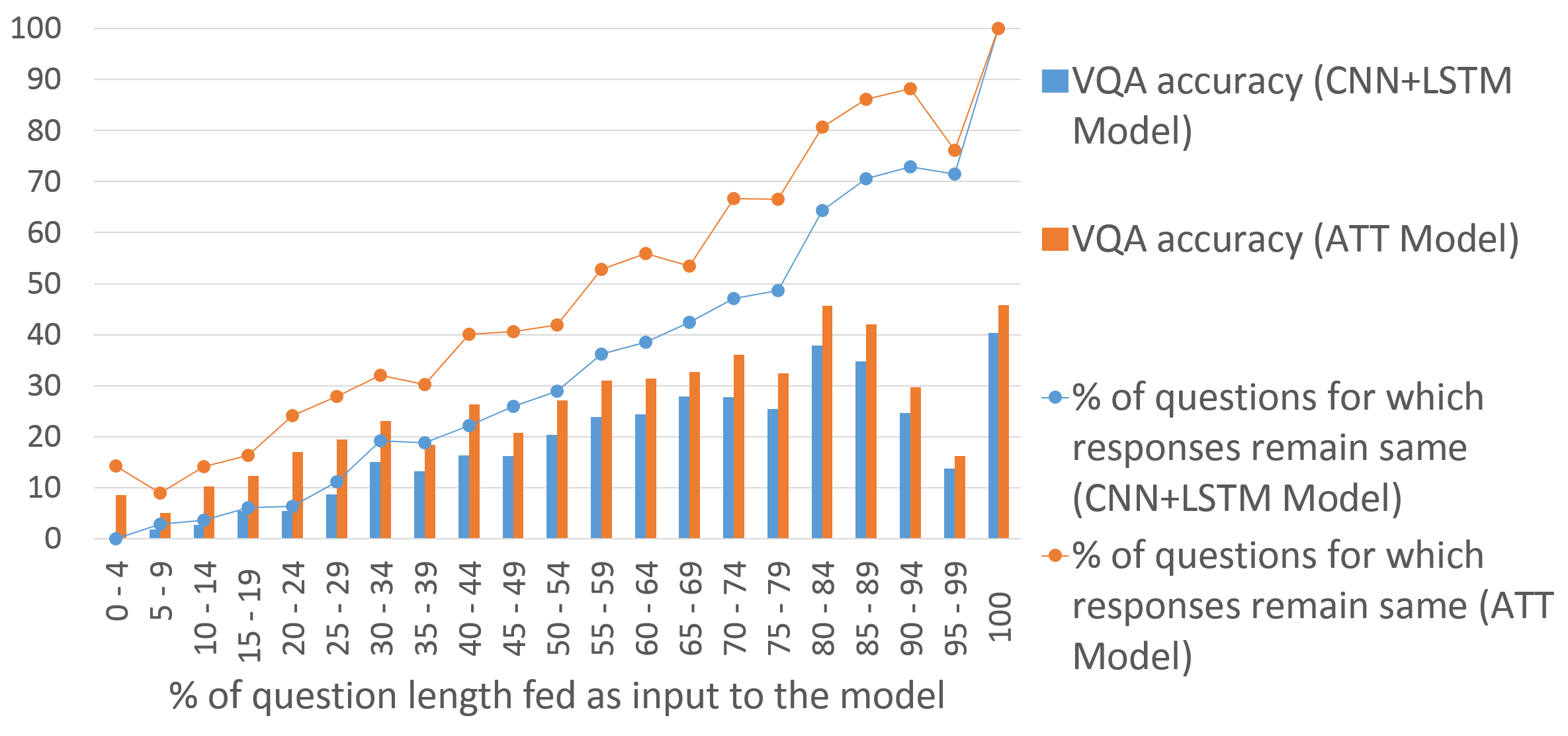}
\caption{X-axis shows length of partial ``other'' question (in \%) fed as input. Y-axis shows percentage of ``other'' questions for which responses of these partial ``other'' questions are the same as full ``other'' questions and VQA accuracy of partial ``other'' questions.}
\label{fig:ques_len_other}
\end{figure}

For ``yes/no'' questions, the \att model seems particularly `jumpy' -- converging on a predicted answer listening to only the first few words of the question (see \figref{fig:ques_len_bin}). Surprisingly, the accuracy is also as much as the final accuracy (after listening to entire question) when making predictions based on first few words of the question. In contrast, the \noatt model converges on a predicted answer later, after listening to atleast 35\% of the question, achieving as much as the final accuracy after convergence. For ``number'' and ``other'' questions, both \att and \noatt model show similar trends (see \figref{fig:ques_len_num} for ``number'' and \figref{fig:ques_len_other} for ``other'').

\begin{figure}[h]
\centering
\includegraphics[width=1\linewidth]{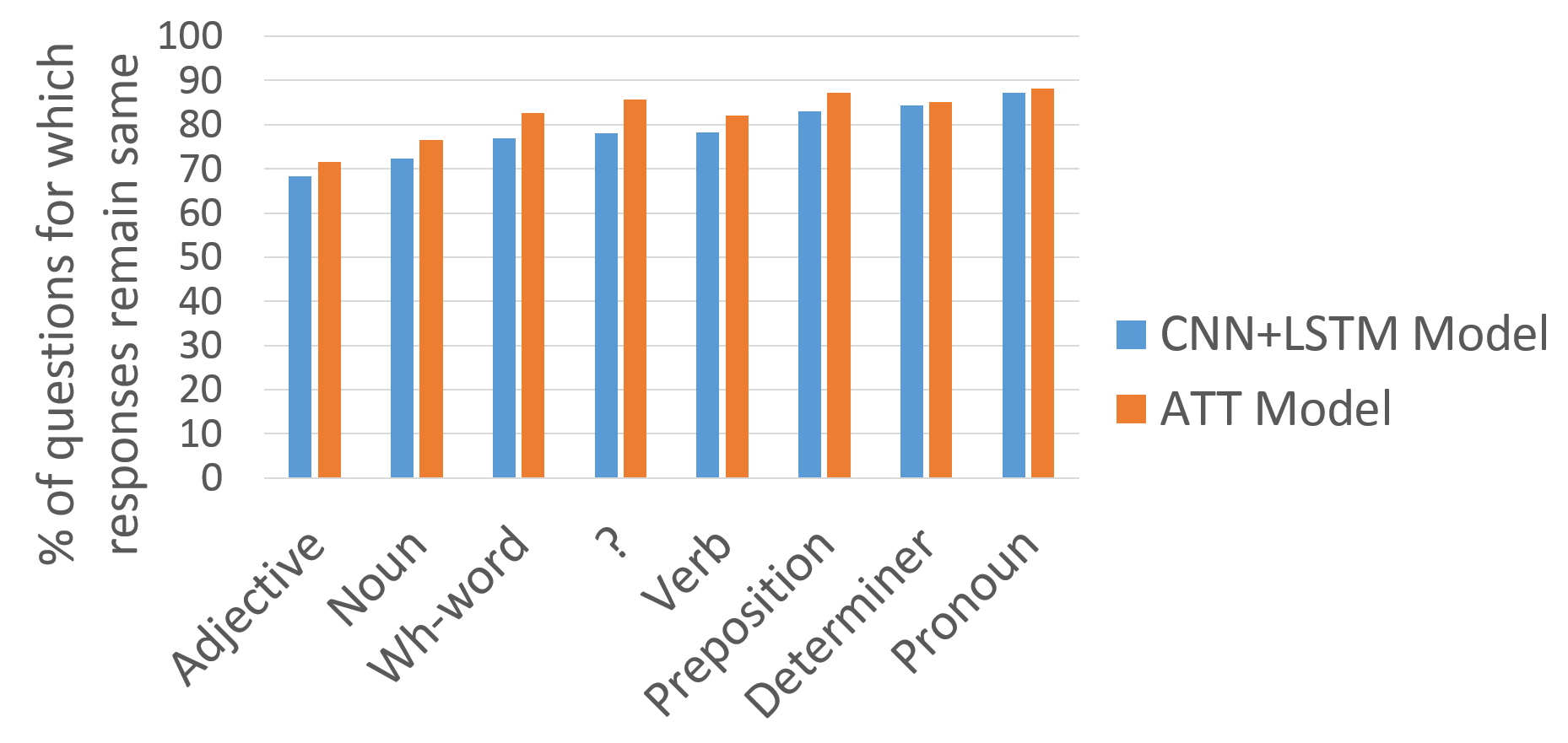}
\caption{Percentage of ``yes/no'' questions for which responses remain same (compared to entire ``yes/no' question) as a function of POS tags dropped from the ``yes/no' question.}
\label{fig:pos_len_bin}
\end{figure}

\begin{figure}[h]
\centering
\includegraphics[width=1\linewidth]{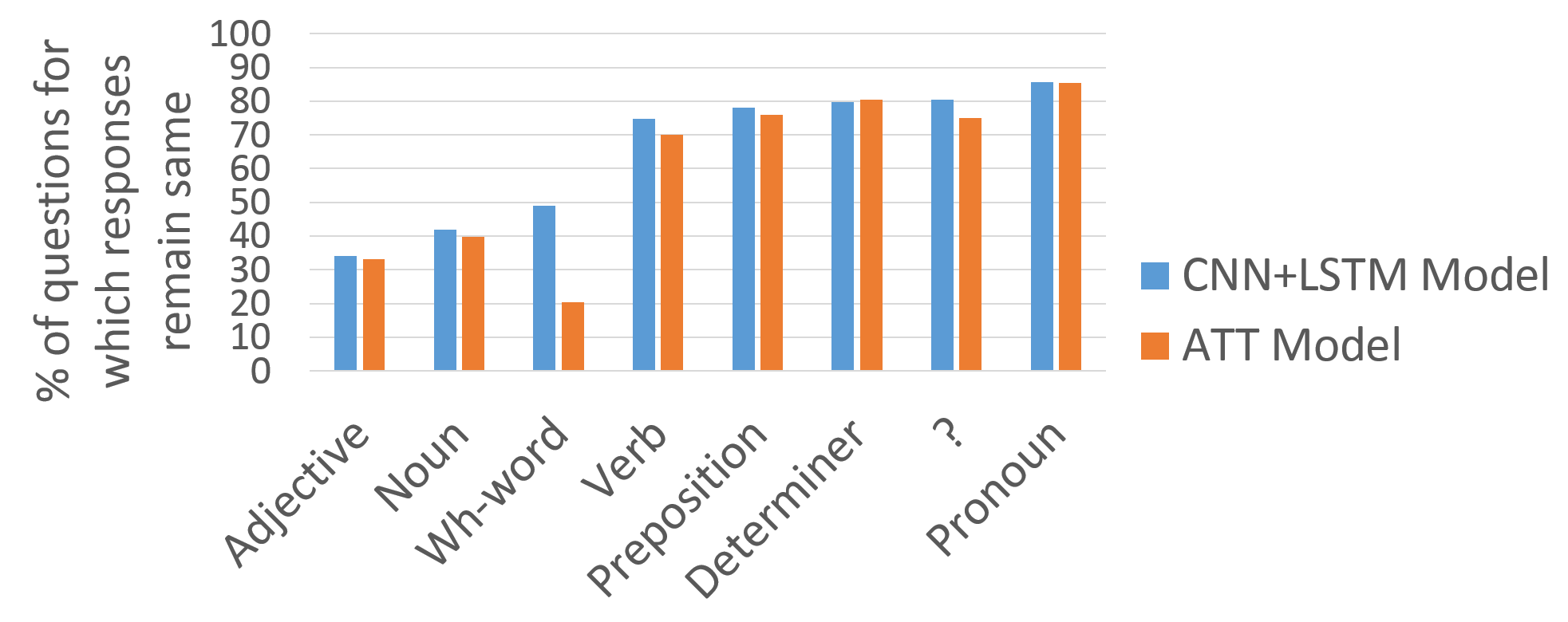}
\caption{Percentage of ``number'' questions for which responses remain same (compared to entire ``number'' question) as a function of POS tags dropped from the ``number'' question.}
\label{fig:pos_len_num}
\end{figure}

\begin{figure}[h]
\centering
\includegraphics[width=1\linewidth]{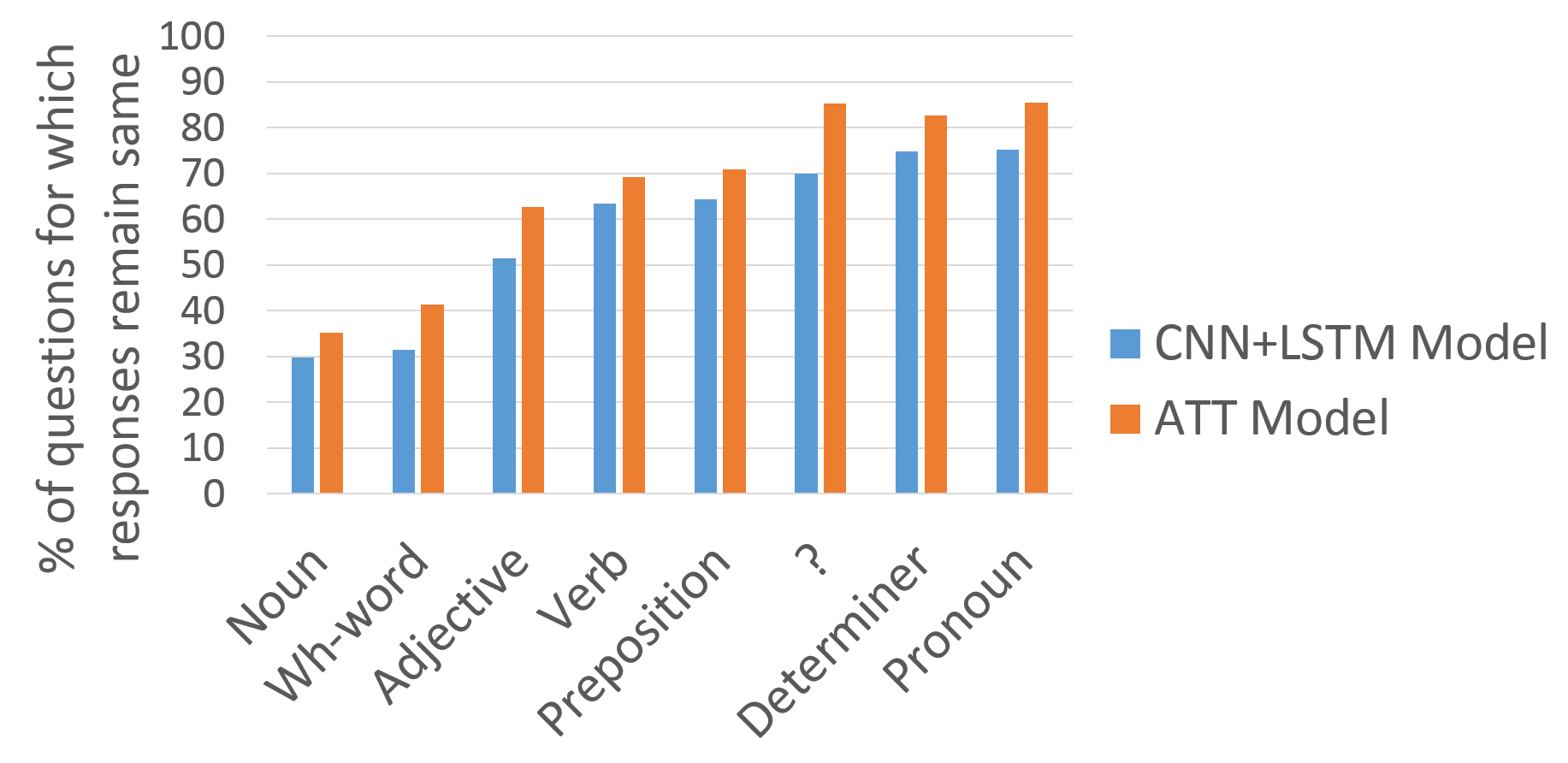}
\caption{Percentage of ``other'' questions for which responses remain same (compared to entire ``other'' question) as a function of POS tags dropped from the ``other'' question.}
\label{fig:pos_len_other}
\end{figure}

It is interesting to note that VQA models are most sensitive to adjectives for ``yes/no'' questions (compared to wh-words for all questions) (see \figref{fig:pos_len_bin}). This is probably because often the ``yes/no'' questions are about attributes of objects (e.g., ``Is the cup empty?''). For ``number'' questions, the \noatt model is most sensitive to adjectives whereas the \att model is most sensitive to wh-words (see \figref{fig:pos_len_num}). For ``other'' questions, both the models are most sensitive to ``nouns'' (see \figref{fig:pos_len_other}).

\section*{Appendix V: Additional qualitative examples for ``complete question understanding''}
\label{sec:sec5}

\begin{figure*}[h]
\centering
\includegraphics[width=1\linewidth]{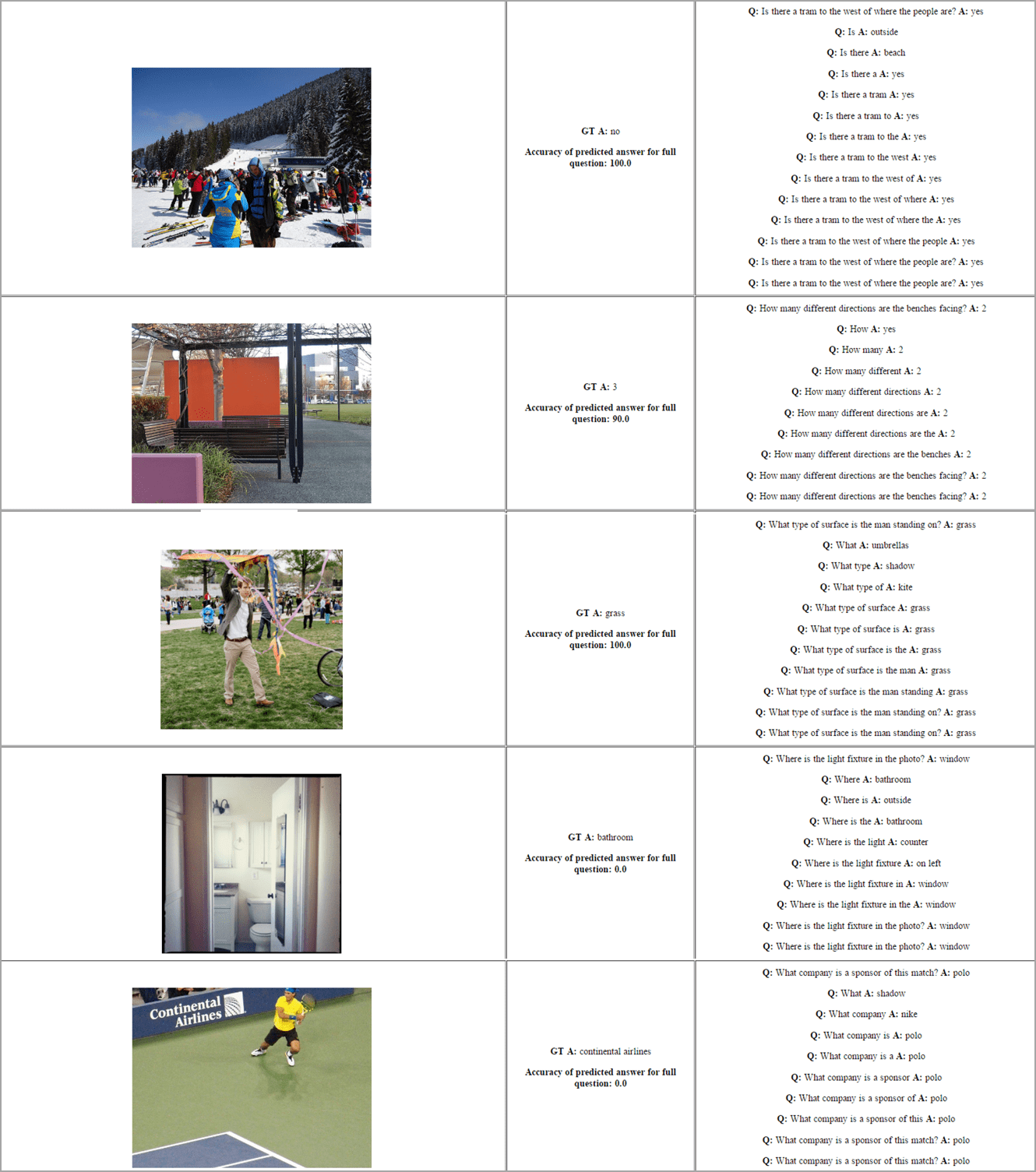}
\caption{Examples where the \noatt model converges on a predicted answer without listening to the entire question.}
\label{fig:qual3_supp}
\end{figure*}

\figref{fig:qual3_supp} shows examples where the \noatt model converges on a predicted answer without listening to the entire question. On doing so, the model gets the answer correct for some QI pairs (first three rows) and incorrect for others (last two rows).

\section*{Appendix VI: Analyses on ``complete image understanding'' for different question types}
\label{sec:sec6}

\begin{figure}[h]
\centering
\includegraphics[width=1\linewidth]{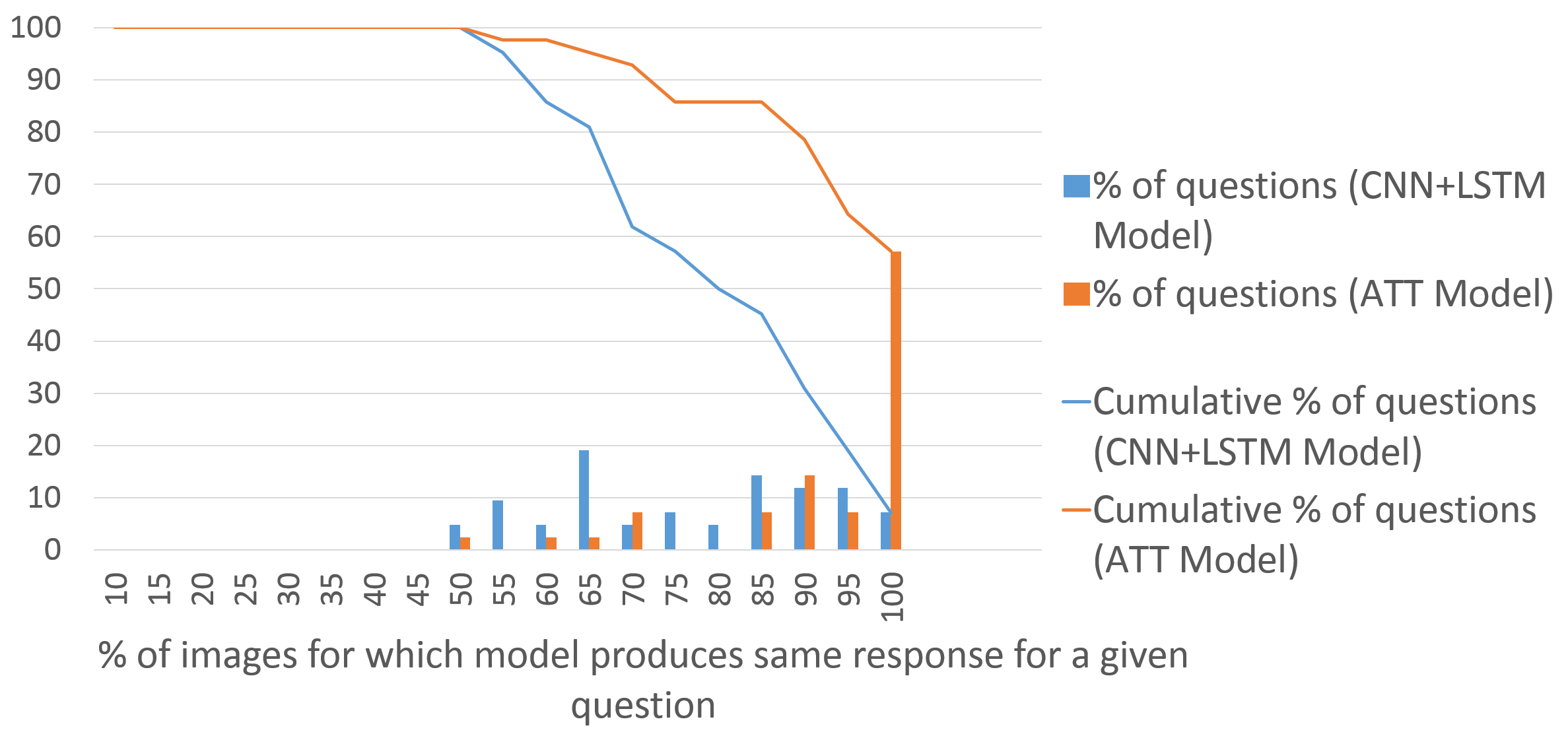}
\caption{Histogram of percentage of images for which model produces same answer for a given ``yes/no'' question. The cumulative plot shows the \% of ``yes/no'' questions for which model produces same answer for \emph{atleast} $x$ \% of images.}
\label{fig:img_und_bin}
\end{figure}

\begin{figure}[h]
\centering
\includegraphics[width=1\linewidth]{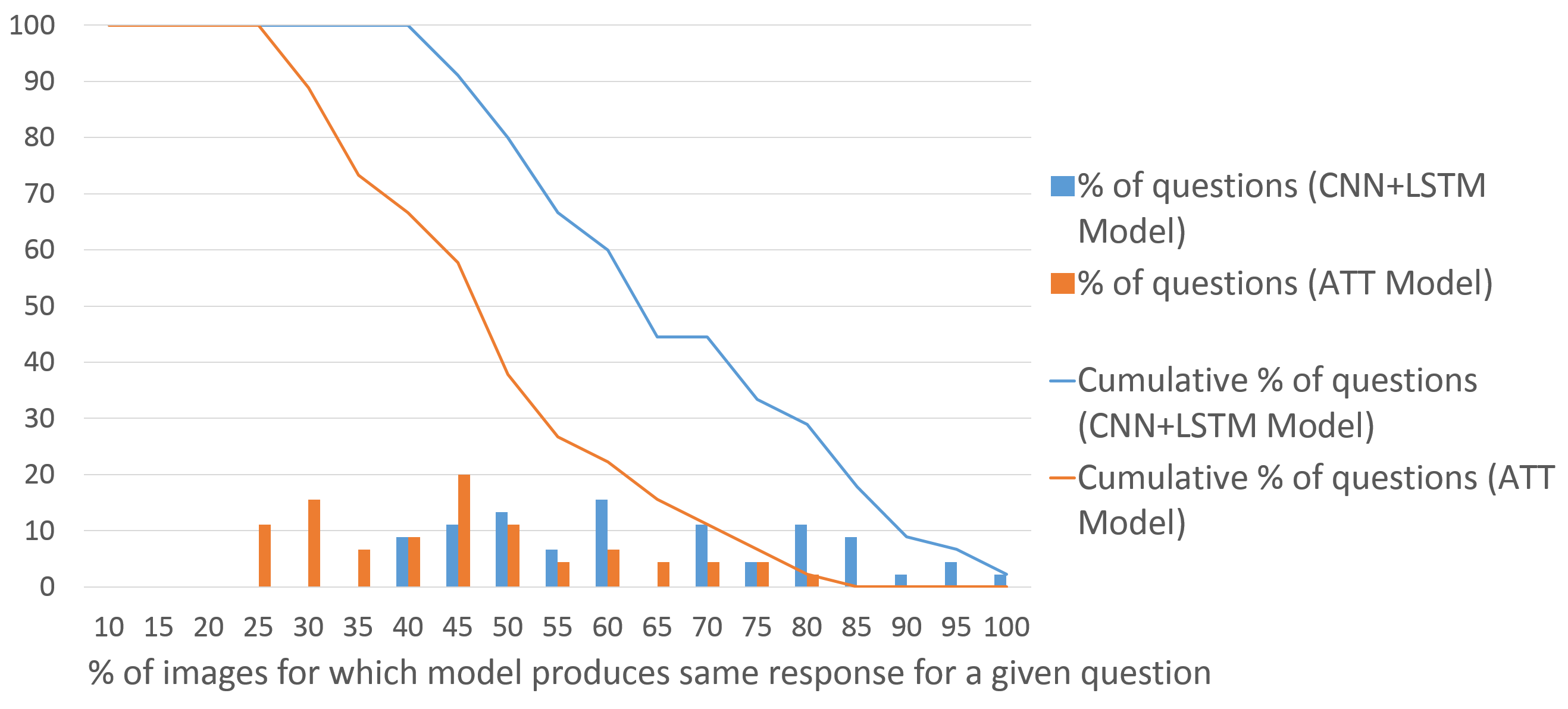}
\caption{Histogram of percentage of images for which model produces same answer for a given ``number'' question. The cumulative plot shows the \% of ``number'' questions for which model produces same answer for \emph{atleast} $x$ \% of images.}
\label{fig:img_und_num}
\end{figure}

\begin{figure}[h]
\centering
\includegraphics[width=1\linewidth]{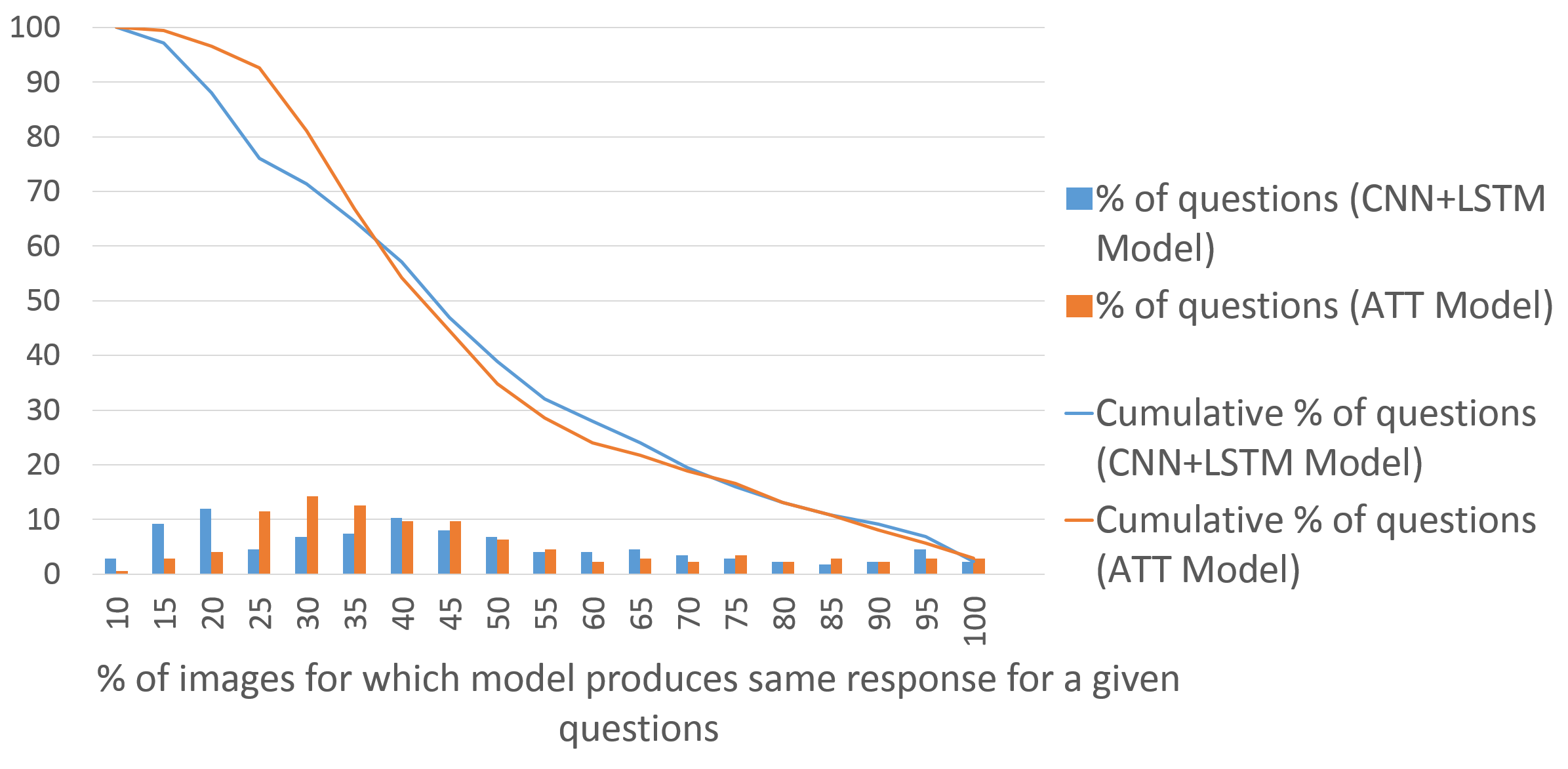}
\caption{Histogram of percentage of images for which model produces same answer for a given ``other'' question. The cumulative plot shows the \% of ``other'' questions for which model produces same answer for \emph{atleast} $x$ \% of images.}
\label{fig:img_und_other}
\end{figure}

\figref{fig:img_und_bin}, \figref{fig:img_und_num} and \figref{fig:img_und_other} show the breakdown of percentage of questions for which the model produces same answer across images for ``yes/no'', ``number'' and ``other'' respectively. The \att model seems to be more ``stubborn'' (does not change its answers across images) for ``yes/no'' questions compared to the \noatt model, and less ``stubborn'' for ``number'' questions compared to the \noatt model. 

\section*{Appendix VII: Additional qualitative examples for ``complete image understanding''}
\label{sec:sec7}

\begin{figure*}[h]
\centering
\includegraphics[width=1\linewidth]{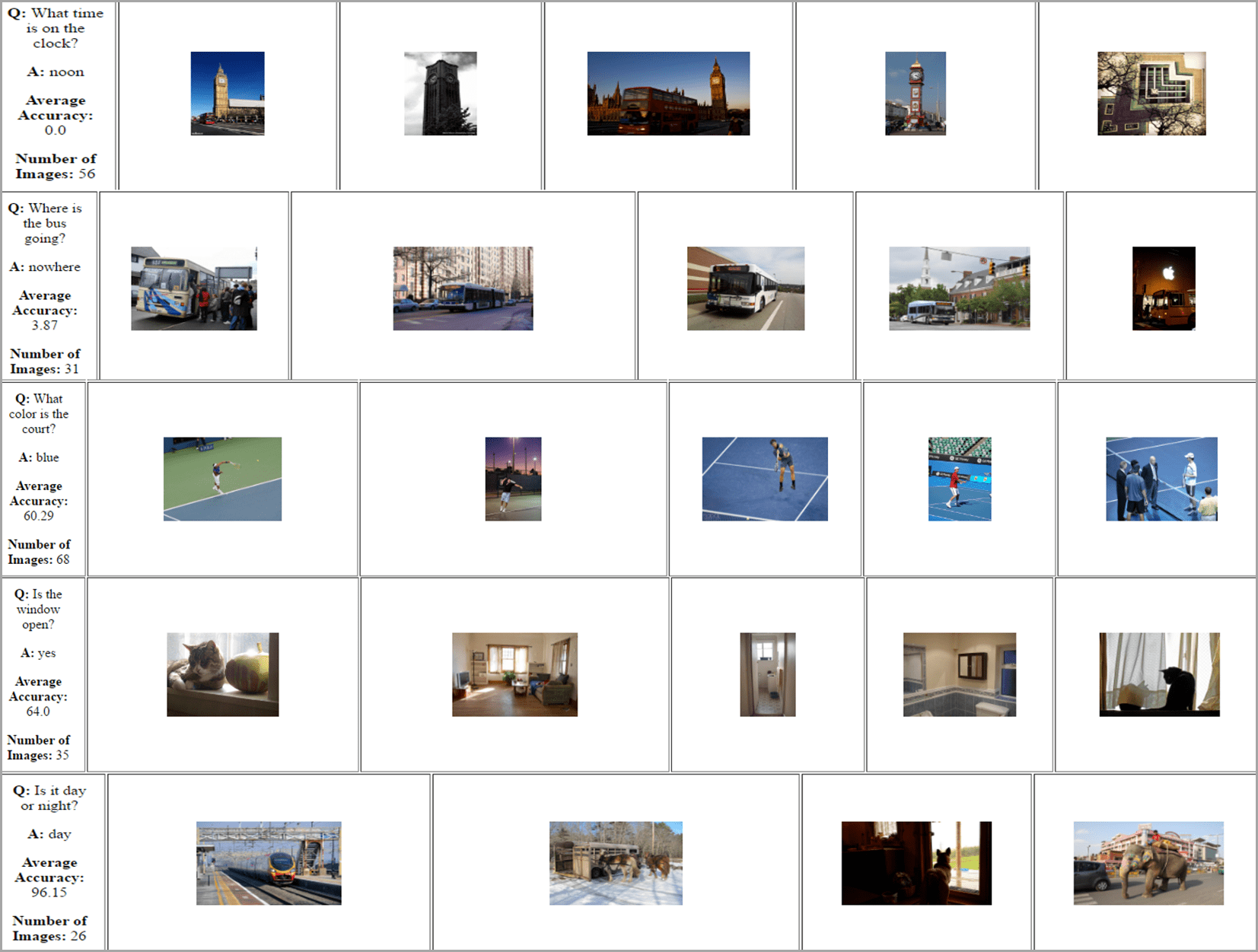}
\caption{Examples where the \noatt model produces the same answer for atleast half the images for each of the questions shown above. ``Q'' denotes the question for which model produces same response for atleast half the images, ``A'' denotes the answer predicted by the model (which is same for atleast half the images), ``Number of Images'' denotes the number of images for which the question is repeated in the VQA validation set and ``Average Accuracy'' is the VQA accuracy for these QI pairs (with same question but different images).}
\label{fig:qual4_supp}
\end{figure*}

\figref{fig:qual4_supp} shows examples where the \noatt model produces the same answer for atleast half the images for a given question and the accuracy achieved by the model for such QI pairs.

